\documentclass{article}
\usepackage{ijcai19}

\usepackage{times}
\usepackage{soul}
\usepackage{url}

\usepackage[utf8]{inputenc}
\usepackage[small]{caption}
\usepackage{graphicx}
\usepackage{amsmath}
\usepackage{booktabs}
\usepackage{bbold}
\usepackage{tabularx,environ,amssymb}
\usepackage{csquotes}
\usepackage{float}
\usepackage{subcaption}
\usepackage{nicefrac}
\usepackage{xcolor}
\newcommand{\alp}[1]{Alp#1}

\newtheorem{definition}{Definition}

\title{Learning Relational Representations with Auto-encoding Logic Programs}

\author{
Sebastijan Dumančić$^1$\footnote{Contact Author}\and
Tias Guns$^2$\and
Wannes Meert$^1$\And
Hendrik Blockeel$^1$\\
\affiliations
$^1$KU Leuven, Belgium\\
$^2$VUB, Belgium\\
\emails
\{sebastijan.dumancic, wannes.meert, hendrik.blockeel\}@cs.kuleuven.be,
tias.guns@vub.be%,
}

\begin{document}

\maketitle

\begin{abstract}
Deep learning methods capable of handling relational data have proliferated over the last years.
In contrast to traditional relational learning methods that leverage first-order logic for representing such data, these deep learning methods aim at re-representing symbolic relational data in Euclidean spaces.
They offer better scalability, but can only numerically approximate relational structures and are less flexible in terms of reasoning tasks supported.
This paper introduces a novel framework for relational representation learning that combines the best of both worlds.
This framework, inspired by the auto-encoding principle, uses first-order logic as a data representation language, and the mapping between the original and latent representation is done by means of logic programs instead of neural networks.
We show how learning can be cast as a constraint optimisation problem for which existing solvers can be used.
The use of logic as a representation language makes the proposed framework more accurate (as the representation is exact, rather than approximate), more flexible, and more interpretable than deep learning methods.
We experimentally show that these latent representations are indeed beneficial in relational learning tasks.\footnote{Supplementary material: \url{https://arxiv.org/abs/1903.12577}}
\end{abstract}

\section{Introduction}

Deep representation learning (DL)~\cite{Goodfellow2016} has proven itself to be an important tool for modern-day machine learning (ML): it \textit{simplifies the learning task} through a series of data transformation steps that define a new feature space (so-called \textit{latent representation}) making data regularities more explicit.
	Yet, DL progress has mainly focused on learning representations for classifiers recognising patterns in sensory data, including computer vision and natural language processing, having a limited impact on representations aiding automated reasoning.
	Learning such reasoning systems falls under the scope of Statistical Relational Learning (SRL)~\cite{Getoor2007}, which combines knowledge representation capabilities of first-order logic with probability theory and hence express both complex relational structures and uncertainty in data.
	The main benefit of SRL models, that most ML methods lack, is the ability to \textit{(1)} operate on any kind of data (feature vectors, graphs, time series) using the same learning and reasoning principles, and \textit{(2)} perform complex chains of reasoning and answer questions about any part of a domain (instead of one pre-defined concept).

	%\seb{the connection between reasoning and relational representations has to be made more explicit}

	%Recognising the need to closely integrate learning and reasoning [put some references], 
	Recent years have yielded various adaptations of standard neural DL models towards reasoning with relational data, namely \textit{Knowledge graph embeddings}~\cite{NickelReview} and \textit{Graph neural networks}~\cite{kipf2017semi,DBLP:journals/debu/HamiltonYL17}.
	These approaches aim to re-represent relational data in vectorised Euclidean spaces, on top of which feature-based machine learning methods can be used.
	Though this offers good learning capabilities, it sacrifices the flexibility of reasoning~\cite{bouchard19} and can only \textit{approximate} relational data, but not capture it in its entirety.

	This work proposes a framework that unites the benefits of both the SRL and the DL research directions. %both benefits -- reasoning of SRL systems and latent representations from DL.
	We start with the question:
	\begin{quote}
	      \textit{Is it possible to learn latent representations of relational data that improve the performance of SRL models, such that the reasoning capabilities are preserved?}
	\end{quote}
	Retaining logic as a representation language for latent representations is crucial in achieving this goal, as retaining it inherits the reasoning capabilities.
	Moreover, it offers additional benefits.
	Logic is easy to understand and interpret (while DL is \textit{black-box}), which is important for trust in AI systems.
	Furthermore, SRL methods allow for incorporation of \textit{expert knowledge} and thus can easily build on previously gathered knowledge.
	Finally, SRL systems are capable of learning from a few examples only, which is in sharp contrast to typically \textit{data-hungry} DL methods.

	We revisit the basic principles of relational representation learning and introduce a novel framework to learn latent representations based on \textit{symbolic}, rather than gradient-based computation.
	The proposed framework implements the \textit{auto-encoder principle}~\cite{Hinton504} -- one of the most versatile deep learning components -- but uses \textit{logic programs} as a computation engine instead of (deep) neural networks.
	For this reason, we name our approach \textit{Auto-encoding logic programs}~(\alp{s}).

	Alongside the formalism of \alp{s}, we contribute a generic procedure to learn \alp{s} from data.
	The procedure translates the learning task to a constraint optimisation problem for which existing efficient solvers can be used.
	In contrast to neural approaches, where the user has to provide the architecture beforehand (e.g., a number of neurons per layer) and tune many hyper-parameters, the output of \alp{s} is an architecture, a relational one, in itself.
	Notably, we show that the learned latent representations help with learning SRL models afterwards: SRL models learned on the latent representation outperform the models learned on the original data representation.

\section{Auto-encoding Logic Programs}

\begin{figure}
	\centering
    \includegraphics[width=.99\linewidth]{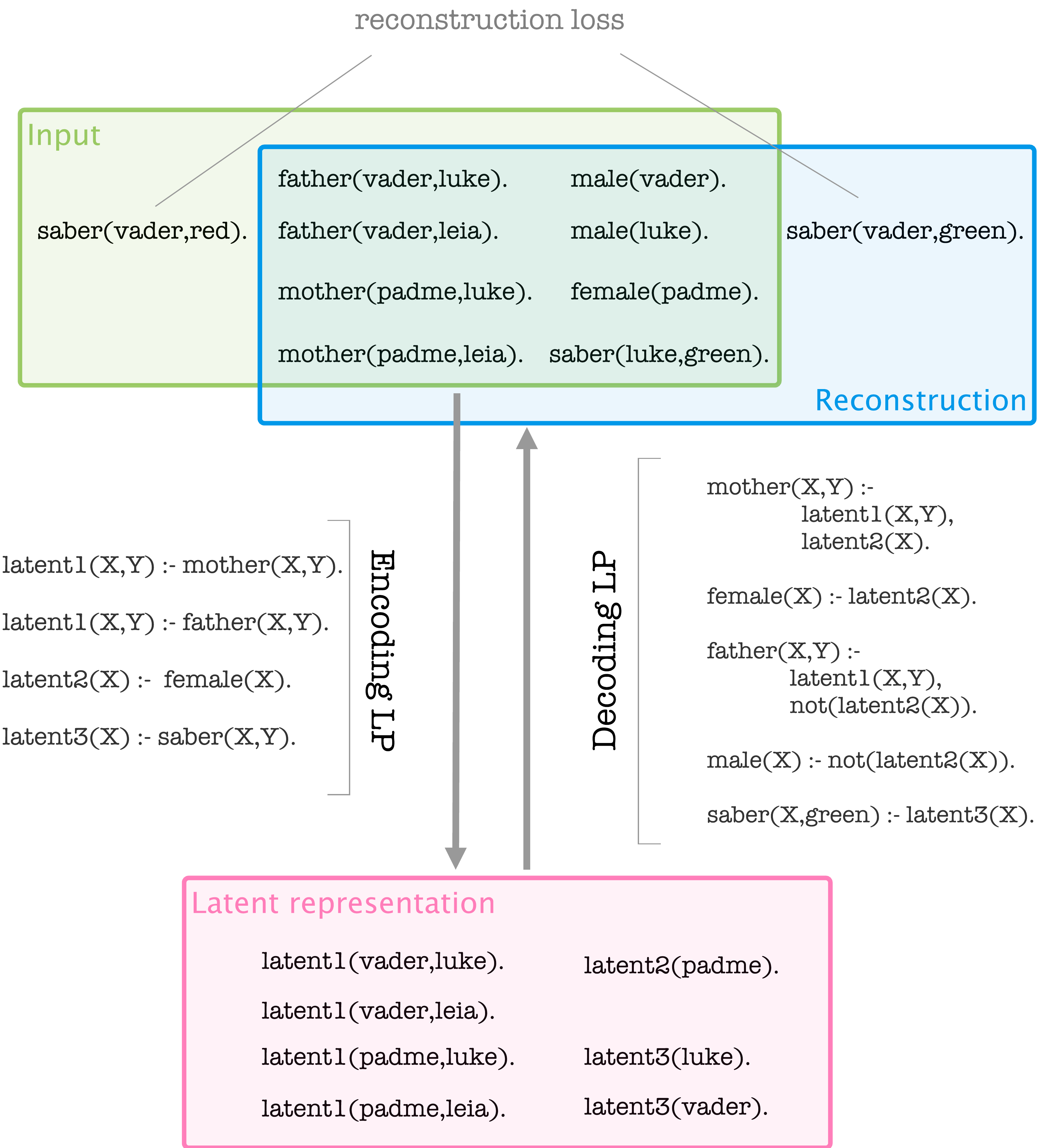}
    \caption{An \textit{auto-encoding logic program} maps the input data, given in a form of a set of facts, to its latent representation through an \textit{encoding logic program}. A \textit{decoding logic program} maps the latent representation of data back to the original data space. The facts missing from the reconstruction (e.g., \textit{saber(vader,red)}) and the wrongly reconstructed facts (e.g., \textit{saber(vader,green)}) consitute the \textit{reconstruction loss}.\label{fig:alp}}
\end{figure}

Auto-encoders learn new representations through the \textit{reconstruction principle}: the goal is to learn an \textit{encoder}, mapping the input data to its latent representation, and a \textit{decoder}, mapping the latent representation back to the original space so that the input data can be faithfully reconstructed.
For a latent representation to be useful, it is important to prevent it from learning an identity mapping -- often done by limiting the dimensionality and/or enforcing sparsity.

In neural auto-encoders, data is represented with vectors and mapping functions are matrices.
Our goal is, intuitively, to lift the framework of auto-encoders to use \textit{first-order logic as a data representation language},  and \textit{logic programs} as mapping functions of encoder and decoder (Figure \ref{fig:alp}).
In the following paragraphs, we describe the basic components of \alp{s}.

\paragraph{Data.}
To handle arbitrary relational data, \alp{s} represent data as a set of logical statements, such as \textit{father(vader,luke)} (Figure \ref{fig:alp}, Input).
These statements consist of \textit{constants} representing the entities in a domain (e.g., \textit{vader}, \textit{luke}) and \textit{predicates} indicating the relationships between entities (e.g., \textit{father}).
A \textit{ground atom} is a predicate symbol applied to constants (e.g., \textit{father(vader,luke)}); if an atom evaluates to \textit{true}, it represents a \textit{fact}.
Given a set of predicates $\mathcal{P}$ and a set of constants $\mathcal{C}$ (briefly, a vocabulary $(\mathcal{P}, \mathcal{C})$ ), the Herbrand base $\mathcal{HB(P,C)}$ is the set of all atoms that can be constructed using $\mathcal{P}$ and $\mathcal{C}$.
A knowledge base is a subset of the Herbrand base; it contains all the atoms that evaluate to \textit{true}.

\vspace{-2pt}

% \textbf{Data.}
% To handle arbitrary relational data, SRL uses a set of logical statements as data representation, such as \textit{father(vader,luke)} (Figure \ref{fig:alp}, Input).
% These statements consist of a \textit{predicate} indicating the relationships between entities (e.g., \textit{father($\cdot$,$\cdot$)}) and \textit{constants} representing the entities in a domain (e.g., \textit{vader}, \textit{luke}).
% %\tias{I added this notation to make arity explicit and predicat more tangible, hopefully arity is in predicate definition, or acceptable as is...}
%
% A \textit{ground atom} is a predicate symbol applied to constants (e.g., \textit{father(vader,luke)}); if an atom evaluates to \textit{true}, it represents a \textit{fact}\tias{evaluating an atom unclear for reader + this is about data, not relevant here? move out evaluation/open-world to relevant part if possible}.
% Thus, the data has a vocabulary consisting of a set of predicates $\mathcal{P}$ and constants $\mathcal{C}$.
% A latent representation shares the same format (see Fig. \ref{fig:alp}, bottom), but relies on a new set of predicates $\mathcal{L}$ with $\mathcal{L} \cap \mathcal{P} = \emptyset$.

\paragraph{Mapping functions.}
The mapping functions of both encoder and decoder are realised as \textit{logic programs}.
A logic program is a set of \textit{clauses} -- logical formulas of the form \textit{h :- b$_1$,\ldots,b$_n$} , where \textit{h} is called the \textit{head} literal and \textit{b$_i$} are \textit{body} literals (comma denotes conjunction).
 A literal is an atom or its negation.
 Literals can contain variables as arguments; these are by definition universally quantified.
 Given a vocabulary $(\mathcal{P}, \mathcal{C})$, we call a literal a $(\mathcal{P}, \mathcal{C})$-literal if its predicate is in $\mathcal{P}$ and its argument are constants in $\mathcal{C}$ or variables.
Clauses are read as logical implications; e.g., the clause \textit{mother(X,Y) :- parent(X,Y),female(X)} states that for all $X$ and $Y$, $X$ is a mother of $Y$ if $X$ is a parent of $Y$ and $X$ is female.

\vspace{-1pt}

% \textbf{Mapping functions.}
% The mapping functions of both encoder and decoder are realised as \textit{logic programs}.
% That is, the mapping is computed through a set of \textit{clauses} -- logical formulas of the form \textit{h :- b$_1$,\ldots,b$_n$}, where \textit{h} is called the \textit{head} atom and \textit{b$_i$} are \textit{body} atoms (comma denotes conjunction \tias{semicolon disjunction}).
% Clauses are read as logical implications: \textit{h} if \textit{b$_1$} and \textit{\ldots} and \textit{b$_n$}.
% %A \textit{logic program} (LP) is a set of clauses.
% An example of a clause is \textit{mother(X,Y) :- parent(X,Y),female(X).}, where symbols like \textit{X} and \textit{Y} stand for \textit{logic variables} ranging over the entities in the domain. \tias{here evaluation? example needed to understand rule. If example clause would be: parent(X,Y) :- mother(X,Y); father(X,Y), then, e.g.: With the input of Fig 1, parent(vader,luke) and parent(padme,luke) would both evaluate to true, while parent(luke,lea) would not.}

\paragraph{Encoding program.}
Given an input vocabulary $(\mathcal{P}, \mathcal{C})$, an \textit{encoding logic program} $\mathcal{E}$ (Fig. \ref{fig:alp} middle left) is a set of clauses with $(\mathcal{P}, \mathcal{C})$-literals in the body and a positive $(\mathcal{L}, \mathcal{C})$-literal in the head, where $\mathcal{L}$ is a set of predicates that is disjoint with $\mathcal{P}$ and is extended by the learner as needed.
$\mathcal{E}$ takes as input a knowledge base $\mathcal{KB} \subseteq \mathcal{HB(P,C)}$ and produces as output a latent representation $\mathcal{KB}’ \subseteq \mathcal{HB(L,C)}$, more specifically the set of all facts that are implied by $\mathcal{E}$ and $\mathcal{KB}$.

\paragraph{Decoding program.}
A \textit{decoding logic program} $\mathcal{D}$ similarly maps a subset of $\mathcal{HB(L,C)}$ back to a subset of $\mathcal{HB(P,C)}$.  Its clauses are termed \textit{decoder clauses}; they contain $(\mathcal{L,C})$ literals in the body and a positive $(\mathcal{P,C})$-literal in the head.

\paragraph{\alp{s}.}
Given encoding and decoding logic programs $\mathcal{E}$ and $\mathcal{D}$, their composition $\mathcal{D} \circ \mathcal{E}$ is called
an \textit{auto-encoding logic program} (\alp{}).
An \alp{} is lossless if for any $\mathcal{KB}$, $\mathcal{D}(\mathcal{E}(\mathcal{KB})) = \mathcal{KB}$.
In this paper, we measure the quality of \alp{s} using the following loss function:

\begin{definition}
\textbf{Knowledge base reconstruction loss.}
The knowledge base reconstruction loss (the disagreement between the input and the reconstruction), $loss(\mathcal{E}, \mathcal{D}, \mathcal{KB})$, is defined as

\begin{equation}
    loss(\mathcal{E},\mathcal{D},\mathcal{KB}) = | \mathcal{D}(\mathcal{E}(\mathcal{KB})) \, \ \  \Delta \, \ \  \mathcal{KB} |
    \label{eq:reconstruction}
\end{equation}

where $\Delta$ is the symmetric difference between two sets.
\end{definition}

% \begin{definition}
% \textbf{Knowledge base reconstruction loss.}
% The knowledge base reconstruction loss (the disagreement between the input and the reconstruction), $loss(\mathcal{E}, \mathcal{D}, \mathcal{KB})$, is defined as
%
% \begin{equation}
%     loss(\mathcal{E},\mathcal{D},\mathcal{KB}) = | \mathcal{D}(\mathcal{E}(\mathcal{KB})) \, \ \  \Delta \, \ \  \mathcal{KB} |
%     \label{eq:reconstruction}
% \end{equation}
%
% where $\Delta$ is the symmetric difference between two sets. % (the given knowledge base $\mathcal{KB}$ and the decoded $\mathcal{D}(\mathcal{E}(\mathcal{KB}))$), which returns (1) all non-reconstructed facts from $\mathcal{KB}$ and (2)  all \textit{false} reconstructions.
% \end{definition}
% To accommodate the open-world assumption, the atoms evaluating to \textit{unknown} are not considered in the reconstruction loss. \tias{move open-world expl here, where needed?}

\section{Learning as Constraint Optimisation}

With the main components of \alp{s} defined in the previous section, we define the learning task as follows:

\begin{definition}
%\textbf{Learning \alp{s}.}
\textbf{Given} a knowledge base $\mathcal{KB}$ and constraints on the latent representation, \textbf{find} $\mathcal{E}$ and $\mathcal{D}$ that minimise $loss(\mathcal{E}, \mathcal{D}, \mathcal{KB})$ and $\mathcal{E}(\mathcal{KB})$ fulfils the constraints.
\end{definition}

The constraints on the latent representation prevent it from learning an identity mapping. For example, enforcing sparsity by requiring that the $\mathcal{E}(\mathcal{KB})$ has at most $N$ facts.
We formally define these constraints later.

\begin{figure}
  \centering
  \includegraphics[width=.95\linewidth]{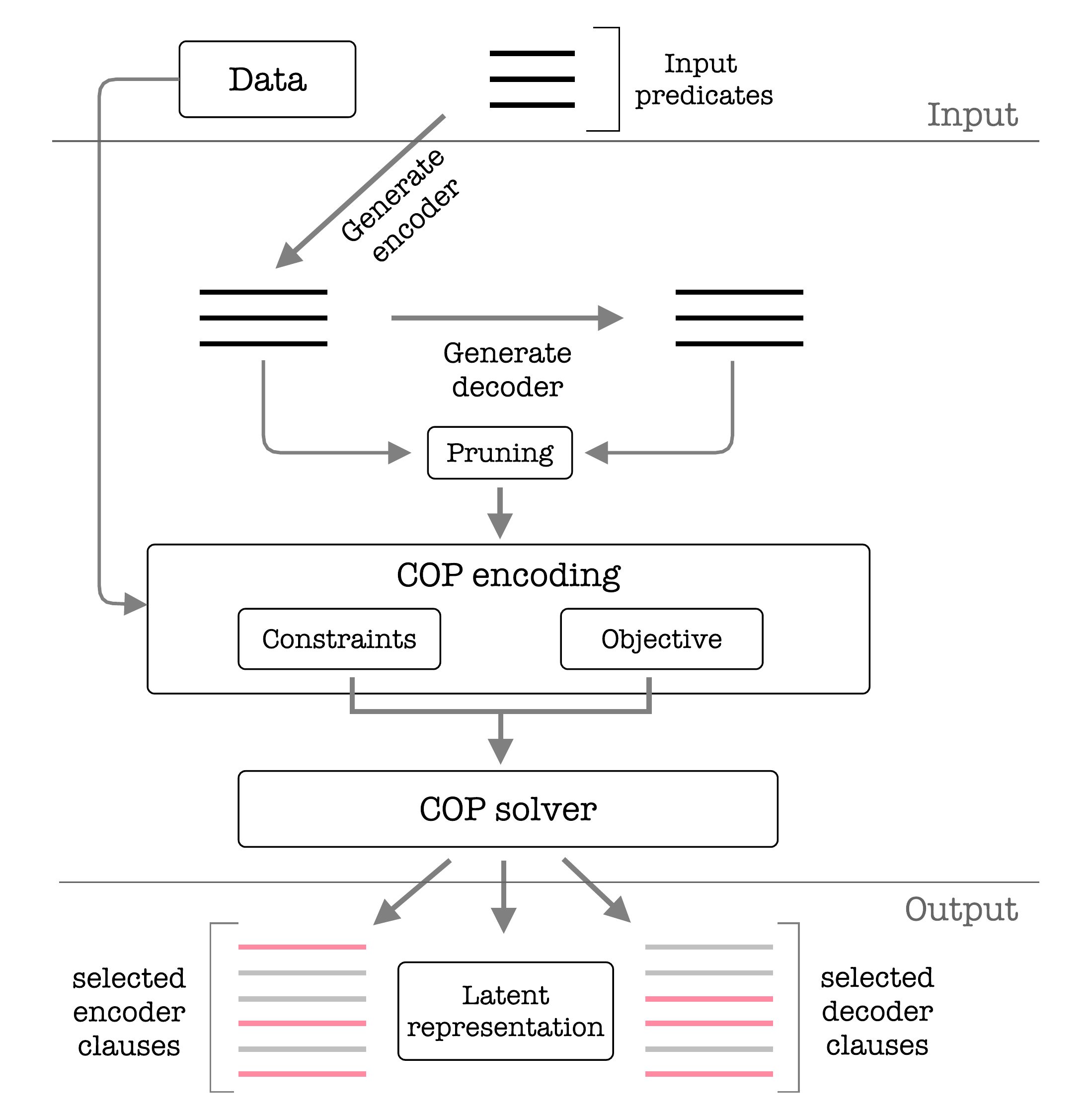}
  \caption{\textit{Learning \alp{s}}. Given the data and a set of predicates as an input, we first enumerate possible encoder clauses and subsequently the decoder clauses. These are used to generate the COP encoding, including the constraints and the objective, which is pruned and passed to the COP solver. The solver returns the selected encoder/decoder clauses and the latent representation.}
  \label{fig:pipeline}
\end{figure}

Intuitively, learning \alp{s} corresponds to \textit{a search for a well-performing combination of encoder and decoder clauses}.
That is, out of a set of possible encoder and decoder clauses, \textit{select} a subset that minimises the reconstruction loss.
To find this subset, we introduce a learning method inspired by the \textit{enumerative} and \textit{constraint solving} techniques from program induction~\cite{Gulwani2017} (illustrated in Figure \ref{fig:pipeline}).
Given a $\mathcal{KB}$ and predicates $\mathcal{P}$, we first enumerate possible encoder clauses.
These clauses define a set of candidate \textit{latent predicates} $\mathcal{L}$ which are subsequently used to generate candidate decoder clauses.
The obtained sets, which define the space of candidate clauses to choose from, are then pruned and used to formulate the learning task as a generic \textit{constraint optimisation problem} (COP) \cite{Rossi:2006:HCP:1207782}.
Such a COP formulation allows us to tackle problems with an extremely large search space and leverage existing efficient solvers. % that can nowadays deal with problems containing millions of variables.
The COP is solved using the Oscar solver\footnote{https://bitbucket.org/oscarlib/oscar/wiki/Home}. The resulting solution is a subset of the candidate encoder and decoder clauses that constitute an \alp{}.

A COP consists of three components: \textbf{decision variables} whose values have to be assigned, \textbf{constraints} on decision variables, and an \textbf{objective function} over the decision variables that expresses the quality of the assignment.
A \textit{solution} consists of a value assignment to the decision variables such that all constraints are satisfied.
In the following sections, we describe each of these components for learning \alp{s}.

\subsection{Decision Variables: Candidate Clauses}

% To generate candidate encoder clauses, we enumerate logical formulas using atoms or their negations, with shared variables up to a certain length.\tias{do we do negation? BTW, watch out with ILP speak here, you used 'literal' though it is not used later in the paper so confusing, and speak here about 'formula' while earlier in text all about clauses}
% The condition of sharing variables forces formulas to describe the same group of objects: for instance, in a formula $p_1(X,Y),p_2(Y,Z)$  a variable $Y$ is shared, while $p_1(X,Y),p_2(Z,W)$ does not contain shared variables.
%
% The enumerated formulas constitute the \textit{bodies} of candidate encoder clauses.
% Before defining a new \textit{latent predicate} that will form the head of the clause, we have to decide which variables from the body to use in the head.
% To do so, we use subsets of variables with the maximal size of the subset equal to the maximum number of arguments of predicates $\mathcal{P}$.
% Each subset is associated with a new latent predicate, which combined for candidate encoder clauses.
% Candidate decoder clauses are generated in the same way, but starting from the predicates $\mathcal{L}$.
%
% Each candidate encoder and decoder clause is associated with a \textit{boolean decision variable} indicating whether a clause is selected (having the value 1) or not (having the value 0).

%\tias{ALt to this entire subsection:}
The COP will have one Boolean decision variable $ec_i$ for each generated candidate encoder clause, and a Boolean decision variable $dc_i$ for each generated candidate decoder clause, indicating whether a clause is selected (having the value 1) or not (having value 0).

To generate the candidate encoder clauses, we start from the predicates in the input data and generate all possible bodies (conjunctions or disjunctions of input predicates with logic variables as entities) up to a given maximum length $l$. Furthermore, we enforce that the predicates share at least one logic variable, e.g. $p_1(X,Y),p_2(Y,Z)$ is allowed while $p_1(X,Y),p_2(Z,W)$ is not.
For each possible body, we then define a new \textit{latent predicate} that will form the head of the clause. This requires deciding which variables from the body to use in the head. We generate all heads that use a subset of variables, with the maximal size of the subset equal to the maximum number of arguments of predicates $\mathcal{P}$.  Candidate decoder clauses are generated in the same way, but starting from the predicates $\mathcal{L}$.

\subsection{Constraints}

\subsubsection{Bottleneck Constraint}
The primary role of constraints in \alp{s} is to impose a bottleneck on the capacity of the latent representation; this is the key ingredient in preventing the auto-encoder from learning the \textit{identity mapping} as $\mathcal{E}$ and $\mathcal{D}$.
This is often done by enforcing compression in the latent representation, sparsity or both.

The straightforward way of imposing compression in \alp{s} is to limit the number of facts in the latent representation.
Preliminary experiments showed this to be a very restrictive setting.
In \alp{s} we impose the bottleneck by limiting the \textit{average number of facts per latent predicate} through the following constraint
$$ \frac{\sum_{i=1}^{N}w_i\text{\texttt{ec}}_i}{\sum_{i=1}^N\text{\texttt{ec}}_i} \leq \gamma G$$
where \texttt{ec}$_i$ are decision variables corresponding to the encoder clauses, $w_i$ is the number of latent facts the encoder clause \texttt{ec}$_i$ entails, $G$ is the average number of facts per predicate in the original data representation and $\gamma$ is the \textit{compression parameter}  specified by the user.
For example, in Figure \ref{fig:alp}, $G=\nicefrac{9}{5}$ and $w=4$ for \textit{latent1(X,Y) :- mother(X,Y); father(X,Y)} .

\subsubsection{Semantic Constraints}

The secondary role of constraints is to impose additional structure to the search space, which can substantially speed up the search.
The following set of constraints reduces the search space by removing undesirable and redundant solutions\footnote{Exact constraint formulations are in the supplementary material}.
These constraints are automatically generated and do not require input from the user.

\paragraph{Connecting encoder and decoder.}
A large part of the search space can be cut out by noticing that the encoder clauses deterministically depend on the decoder clauses.
For instance, if a decoder clause \textit{mother(X,Y) :- latent1(X,Y),latent2(X)} is selected in the solution, then the encoder clauses defining the latent predicates \textit{latent1} and \textit{latent2}  have to be selected as well.
Consequently, encoder clauses are implied by decoded clauses and search only has to happen over candidate decoder clauses.
The implication is modelled with a constraint ensuring that \textit{the final solution must contain an encoder clause defining a predicate \texttt{l} if the solution contains at least one of the decoder clauses that use \texttt{l} in the body}.

% constraints of the following two types:
% \begin{center}
% 	\texttt{dc}$_k$ $\Rightarrow$ \texttt{ec$_i$ $\wedge$  ec$_j$}
%
%     \texttt{ec}$_i$ $\Rightarrow$ \texttt{dc$_m$ $\vee$ dc$_n$}
% \end{center}
% The first constraint states the if a decoder clause \texttt{dc}$_k$ is selected, then encoder clauses \texttt{ec$_i$} and \texttt{ec$_j$} have to be selected as well (assume they are used in the body of \texttt{dc}$_k$).
% The second constraint state that an encoder clause \texttt{l}$_i$ can be selected only if at least one of the decoder clauses using it in the body, \texttt{dc$_m$} and \texttt{dc$_n$}, is selected.

\paragraph{Generality.}
Given the limited capacity of the latent representation, it is desirable to prevent the solver from ever exploring regions where clauses are too similar and thus yielding a marginal gain.
One way to establish the similarity of clauses is to analyse the ground atoms the clauses cover: a clause $c_1$ is said to be more \textit{general} than a clause $c_2$ if all examples entailed by $c_2$ are also entailed by $c_1$.
As $c_2$ cannot bring new information if $c_1$ is already a part of the solution, we introduce constraints ensuring that \textit{if a clause \texttt{c}$_1$ is more general than a clause \texttt{c}$_2$, at most one of them can be selected}. %\tias{This is stricter than 'if the general one, not the more specific one', what you write is what you do?}

\paragraph{Reconstruct one of each input predicates.}
If $\mathcal{KB}$ contains a predicate with a substantially larger number of facts than the other predicates in $\mathcal{KB}$, a trivial but undesirable solution is one that focuses on reconstructing the predicate and its facts while ignoring the predicates with a smaller number of facts.
To prevent this, we impose the constraints ensuring that \textit{among all decoder clauses with the same input predicate in the head, at least one has to be a part of the solution}. This, of course, does not mean all facts of each input predicate will be reconstructed.
We did notice that this constraint allows the solver to find a good solution substantially faster.

\subsection{Objective Function: The Reconstruction Loss}

We wish to formulate the objective over all \textit{missing} (in $\mathcal{KB}$ but not being reconstructed) and \textit{false reconstructions} (produced by the decoder, but not in $\mathcal{KB}$).
To do so, we first obtain a union of \textit{latent facts} generated by each of the candidate encoder clauses; these are a subset of $\mathcal{HB}(\mathcal{L},\mathcal{C})$.
These latent facts allow us to obtain a union of all ground atoms generated by the candidate decoder clauses; these form a \textit{reconstruction} and are a subset of $\mathcal{HB}(\mathcal{P},\mathcal{C})$.
Additionally, for each ground atom in the reconstruction, we remember which candidate decoder clause reconstructed it.

We hence use the above correspondence between the candidate decoder clauses and the reconstructions to create an auxiliary Boolean decision variable \texttt{rf}$_i$ for each possible ground atom in $\mathcal{HB}(\mathcal{P},\mathcal{C})$ that can be reconstructed.
Whether it is reconstructed or not depends on the decoder clauses that are in the solution.
% The key insight in incorporating the objective into the COP is to notice that, as we have access to all candidate clauses, we can find out which ground atoms they cover.
% To do so, we first evaluate candidate encoder clauses; this gives us the set of possible \textit{latent facts} that could be defined by the encoder clauses.
% Using these latent facts, we evaluate the candidate decoder clauses to obtain the set of possible reconstructions.
% Additionally, for each ground atom in the reconstruction, we remember which candidate decoder clause reconstructed it.

% We wish to formulate the objective over all (wrongly) reconstructed facts. We hence use the above dependency to create an auxiliary Boolean decision variable $rf_i$ for each possible ground atom in $\mathcal{HB}(\mathcal{P},\mathcal{C})$ that can be reconstructed. Whether it is reconstructed or not depends on the decoder clauses that are in the solution.

For example, assume that \textit{mother(padme,leia)} can be reconstructed with either of the following decoder clauses:

\vspace{-14pt}
\begin{align*}
  \footnotesize
  mother(X,Y) \, & \text{:-} \, latent1(X,Y),latent2(X). \\
  mother(X,Y) \,  & \text{:-} \, latent3(X,Y).
\end{align*}
Let the two decoder clauses correspond to the decision variables \texttt{dc}$_1$ and \texttt{dc}$_2$. We introduce \texttt{rf}$_i$ to represent the reconstruction of fact \textit{mother(padme,leia)} and add a constraint
$$ \text{\texttt{rf}}_i \Leftrightarrow  \text{\texttt{dc}}_1  \vee \text{\texttt{dc}}_2. $$

Associating such boolean variable \texttt{rf}$_{e}$ with every $e \in \mathcal{HB}(\mathcal{P},\mathcal{C})$, we can formulate the objective as

\begin{equation}
 \text{minimize} \underbrace{\sum_{i \in \mathcal{KB}}\text{not}(\text{\texttt{rf}}_i)}_{\text{\textit{missing} reconstruction}} +  \underbrace{\sum_{j \in \mathcal{HB}(\mathcal{P}, \mathcal{C}) \setminus \mathcal{KB}} \text{\texttt{rf}}_j.}_{\text{\textit{false} reconstruction}}
\end{equation}

\vspace{-5pt}

\subsection{Search}

Given the combinatorial nature of \alp{s}, finding the optimal solution exactly is impossible in all but the smallest problem instances.
Therefore, we resort to the more scalable technique of \textit{large neighbourhood search} (LNS) \cite{Ahuja:2002:SVL:772382.772385}.
LNS is an iterative search procedure that, in each iteration, performs the exact search over a subset of decision variables.
This subset of variables is called the \textit{neighbourhood} and it is constructed around the best solution found in the previous iterations.

A key design choice in LNS is the construction of the neighbourhood.
The key insight of our strategy is that the solution is necessarily sparse -- only a tiny proportion of candidate decoder clauses will constitute the solution at any time.
Therefore, it is important to preserve at least some of the selected decoder clauses between the iterations.
Let a variable be \textit{active} if it is part of the best solution found so far, and \textit{inactive} otherwise.
We construct the neighbourhood by remembering the value assignment of $\alpha$ \% active variables (corresponding to decoder clauses), and  $\beta$ \% inactive variables corresponding to encoder clauses.
For the individual search runs, we use \textit{last conflict search} \cite{COS} and the \textit{max degree} ordering of decision variables.

\vspace{-4pt}

\subsection{Pruning the Candidates}
As the candidate clauses are generated naively, many candidates will be uninformative and introduce mostly false reconstructions.
It is therefore important to help the search by pruning the set of candidates in an insightful and non-trivial way.
We introduce the following three strategies that leverage the specific properties of the problem at hand.
\vspace{-1pt}

\paragraph{Naming variants.}
%Consider two encoder clauses.
Two encoder clauses are \textit{naming variants} if and only if they reconstructed the same set of ground atoms, apart from the name of the predicate of these ground atoms.
As such clauses contain the same information w.r.t. the constants they contain, we detect all naming variants and keep only one instance as a candidate.
\vspace{-1pt}

\paragraph{Signature variants.}
%Consider two decoder clauses having the same predicate in the head atom.
Two decoder clauses are \textit{signature variants} if and only if they reconstructed the same set of ground atoms and their bodies contain the same predicates.
As signature variants are redundant w.r.t. the optimisation problem, we keep only one of the clauses detected to be signature variants and remove the rest.
\vspace{-1pt}

\paragraph{Corruption level.}
We define the \textit{corruption level} of a decoder clause as a \textit{proportion of the false reconstructions in the ground atoms reconstructed by the decoder clause}.
This turns out to be an important notion: if the corruption level of a decoder clause is greater than $0.5$ then the decoder clause cannot improve the objective function as it introduces more \textit{false} than \textit{true reconstructions}. % \tias{Not entirely true, the objective function is the loss which only counts the negatives... I understand you mean accuracy, but that is not the objective}
We remove the candidate clauses that have a corruption level $\geq 0.5$.
\vspace{-1pt}

These strategies are very effective: applying all three of them during the experiments has cut out more than 50 \% of candidate clauses.

\section{Experiments and Results}
\label{sec:expr}

\begin{figure*}
    \centering
    \includegraphics[width=.99\linewidth]{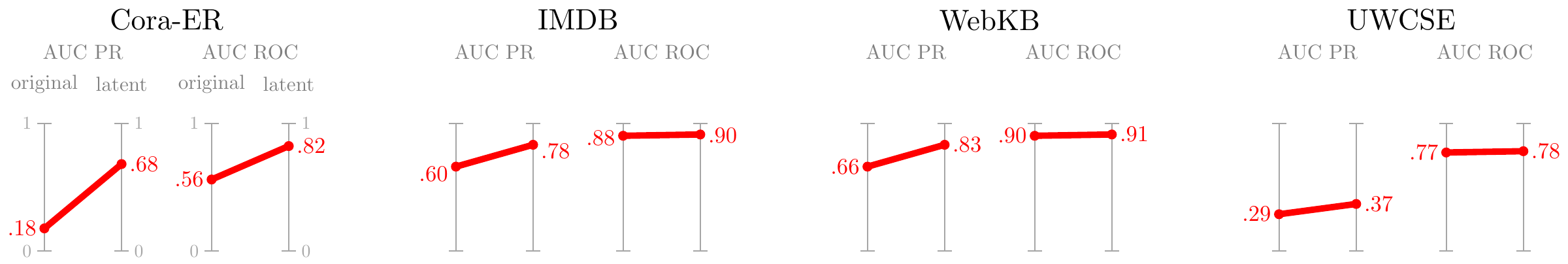}
    \caption{The MLN models learned on the latent data representations created by \alp{s} outperform the MLN models learned on the original data representation, in terms of the AUC-PR scores (red line indicate the increase in the performance), on all dataset. The AUC-ROC scores, which are less reliable due to the sensitivity to class imbalance, remain unchanged.}
    \label{fig:resultsAUCPR}
    \vspace{-12pt}
\end{figure*}

The experiments aim at answering the following question:

\begin{displayquote}
\textbf{Q:} \textit{Does learning from latent representations created by \alp{s} improve the performance of an SRL model?}
\end{displayquote}

We focus on learning generative SRL models, specifically generative \textit{Markov Logic Networks} (MLN) \cite{Richardson:2006:MLN:1113907.1113910}.
The task of generative learning consists of learning a single model capable of answering queries about any part of a domain (i.e., any predicate).
Learning an SRL model consists of searching for a set of logical formulas that will be used to answer the queries.
Therefore, we are interested in whether learning the structure of a generative model in \textit{latent space}, and decoding it back to the original space, is more effective  than learning the model in the original data space.

We focus on this task primarily because no other representation learning method can address this task.
For instance, embeddings vectorise the relational data and thus cannot capture the generative process behind it, nor do they support conditioning on evidence.

The deterministic logical mapping of \alp{s} might seem in contrast with the probabilistic relational approaches of SRL.
However, that is not the case as the majority of SRL approaches consider data to be deterministic and express the uncertainty through the probabilistic model.

\paragraph{Procedure.}
We divide the data in training, validation and test sets respecting the originally provided splits.
The models are learned on the training set, their hyper-parameters tuned on the validation set (in the case of \alp{s}) and tested on the test set.
This evaluation procedure is standard in DL, as full cross-validation is infeasible.
We report both AUC-PR and AUC-ROC results for completeness; note, however, that the AUC-PR is the more relevant measure as it is less sensitive to class imbalance~\cite{Davis:2006:RPR:1143844.1143874}, which is the case with the datasets we use in the experiments.
We evaluate the MLNs in a standard way: we query facts regarding one specific predicate given everything else as evidence and repeat it for each predicate in the test interpretation.

\paragraph{Models.}
We are interested in whether we can obtain better SRL models by learning from the latent data representation.
Therefore, we compare the performance of an MLN learned on the original representation (the \textbf{baseline} MLN) and an MLN learned on the latent representation (the \textbf{latent} MLN) resulting from \alp{s}.
To allow the comparison between the latent and the baseline MLNs, once the latent MLN is learned we add the corresponding decoder clauses as deterministic rules.
This ensures that the baseline and latent MLNs operate in the same space when being evaluated.
\vspace{-1pt}

\paragraph{Learner.}
Both the baseline and the latent MLNs are obtained by the BUSL learner~\cite{mihalkova:icml07}.
We have experimented with more recent MLN learner LSM~\cite{Kok:2010:LML}, but tuning its hyper-parameters proved challenging and we could not get reliable results.
Note that our main contribution is a method for learning \alp{s} and subsequently the latent representation of data, not the structure of an MLN; MLNs are learned on the latent representation created by \alp{s}.
Therefore, the exact choice of an MLN learner is not important, but whether latent representation enables the learner to learn a better model is.
\vspace{-1pt}

\paragraph{Practical considerations.}
We limit the expressivity of MLN models to formulas of length 3 with at most 3 variables (also known as a liftable class of MLNs).
 %the reasoning being that the benefit of the latent representation is easier to measure if the model being learned is simple and
This does not sacrifice the predictive performance of MLNs, as shown by \citeauthor{VanHaaren2016} [\citeyear{VanHaaren2016}].
Imposing this restriction allows us to better quantify the contribution of latent representations: \textit{given a restricted language of the same complexity if the latent MLN performs better that is clear evidence of the benefit of latent representations}. % which achieves the state-of-the-art structure learning results while imposing the same restrictions to MLNs.
The important difference when performing inference with a latent MLN is that each latent predicate that could have been affected by the  removal of the test predicate (i.e., the test predicate is present in the body of the encoder clause defining the specific latent predicate). Hence it has to be declared \textit{open world}, otherwise, MLNs will assume that all atoms not present in the database are \textit{false}.
\vspace{-1pt}

\paragraph{\alp{s} hyper-parameters.}
As with standard auto-encoders, the hyper-parameters of \alp{s} allow a user to tune the latent representation to its needs.
To this end, the hyper-parameters pose a trade-off between the expressivity and efficiency.
When learning latent representations, we vary the length of the encoder and decoder clauses separately in $\{2,3\}$ and the compression level (the $\alpha$ parameter) in $\{0.3, 0.5, 0.7\}$.
\vspace{-1pt}

%If more complexity is needed, several \alp{s} can be learned as a hierarchy.\seb{better keep out?}
%This choice of parameters expresses a trade-off between long, complex rules and the search space size.
%Also, this does not exclude learning longer rules since several layers of latent predicates \alp{s} can be \textit{stacked}.
% This is a useful property in an SRL setting.
% Suppose a formula of length 6 is ideal.
% Instead of searching in a huge space to find such a formula, the latent representation captures short formulas and stacks those to build the longer formula of size 6.
% \seb{this is too confusing and unnecessary?}

\paragraph{Data.}
We use standard SRL benchmark datasets often used with MLN learners: Cora-ER, WebKB, UWCSE and IMDB.
The descriptions of the datasets are available in \cite{mihalkova:icml07,Kok:2010:LML}, while the datasets are available on the Alchemy website\footnote{http://alchemy.cs.washington.edu/}.
% We use standard SRL benchmark datasets often used with MLN learners~\cite{mihalkova:icml07,Kok:2010:LML}.
% \textit{WebKB} contains a set of web pages scrapped from four US universities - Cornell and Universities of Texas, Washington and Wisconsin.
% The data contains web pages, their mutual links and more semantic information about relationships between pages, such as whether the page belongs to a student or a faculty member and courses they teach (we omit the word information, as in \cite{mihalkova:icml07}).
% \textit{Cora-ER} contains a set of papers, their authors, title and venues, together entity resolution information (i.e., titles, authors and venues that refer to the same entities but are spelled differently).
% \textit{UWCSE} contains the information about the employees of five departments at the CS department of University of Washington: student and professors together with mentorship relationships, courses they teach and so on.
% \textit{IMDB} is a small snapshot of the IMDB database describing the relationships between actors, directors and movies.

\subsection{Results}

The results (Figure~\ref{fig:resultsAUCPR}) indicate that BUSL is able to learn better models from the latent representations.
We observe an improved performance, in terms of the AUC-PR score, of the latent MLN on all datasets.
The biggest improvement is observed on the Cora-ER dataset: the latent MLN achieves a score of 0.68, whereas the baseline MLN achieves a score of 0.18.
The IMDB and WebKB datasets experience smaller but still considerable improvements: the latent MLNs improve the AUC-PR scores by approximately 0.18 points.
Finally, a more moderate improvement is observed on the UWCSE dataset: the latent MLN improves the performance for 0.09 points.

These results indicate that latent representations are a useful tool for relational learning.
The latent predicates capture the data dependencies more explicitly than the original data representation and thus can, potentially largely, improve the performance.
This is most evident on the Cora-ER dataset.
To successfully solve the task, a learner has to identify complex dependencies such as \textit{two publications that have a similar title, the same authors and are published at the same venue are identical}.
Such complex clauses are impossible to express with only three predicates; consequently, the baseline MLN achieves a score of 0.18.
However, the latent representation makes these pattern more explicit and the latent MLN performs much better, achieving the score of 0.68.
%\tias{Another thing is that you allow DISjunction and negation, while learners are conjunctive only (perhaps?) hence increasing expressivity again? Could be a property to highlight...}

Neural representation learning methods are sensitive to the hyper-parameter setup, which tend to be domain dependent.
We have noticed similar behaviour with \alp{s} by inspecting the performance on the validation set (details in the supplement).
The optimal parameters can be selected, as we have shown,  on a validation set with a rather small grid as \alp{s} have only three hyper-parameters.

% It is interesting to see that there is no general trend relating reconstruction loss and performance (Figure \ref{fig:lossvsaucpr}).
% On the WebKB, the best performing models are those with (near) perfect reconstruction; on the Cora-ER, the best performing models seems to have relatively high reconstruction loss; while on the UWCSE, the best performing models seem to occupy both parts of the spectrum.
% This might indicate the amount of noise present in the data, suggesting that WebKB is a rather clean dataset, while Cora-ER contains quite some noise and the best performing models are learned on latent representation that manage to reduce the amount of noise.

\begin{figure}
	\centering
	% \begin{subfigure}[t]{0.45\linewidth}
	% 	\centering
	% 	\includegraphics[width=.6\linewidth]{images/lossVsAUC}
	% 	\caption{Relationship between reconstruction loss and AUC-PR scores. \label{fig:lossvsaucpr}}
	% \end{subfigure}
	% \hspace{.5em}
	% \begin{subfigure}[t]{0.45\linewidth}
		\centering
		\includegraphics[width=.83\linewidth]{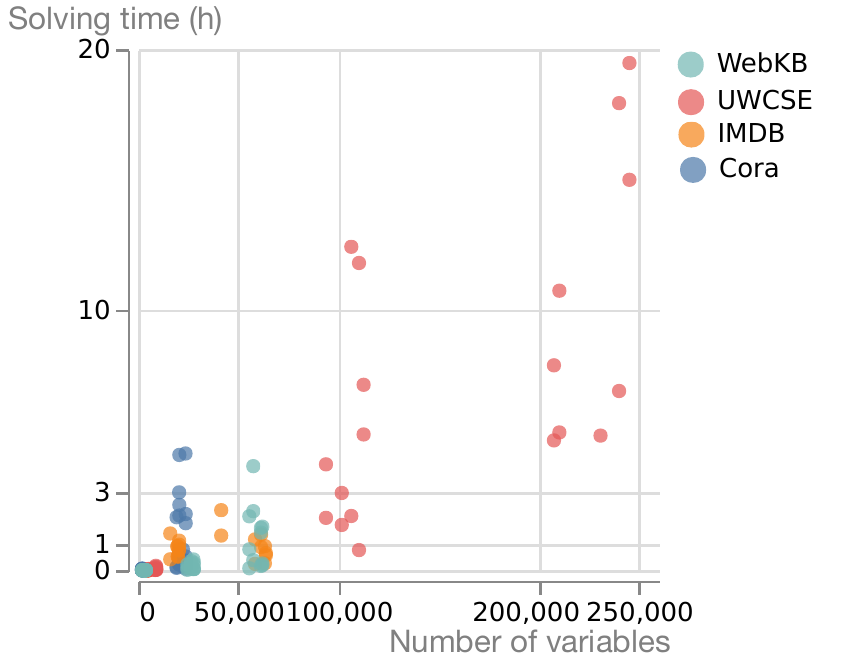}

		\caption{Relationship between runtimes and the number of variables.\label{fig:timings}}
	% \end{subfigure}
\end{figure}

\paragraph{Runtime.}
Figure \ref{fig:timings} summarises the time needed for learning a  latent representation.
These timings show that, despite their combinatorial nature, \alp{s} are quite efficient: the majority of latent representations is learned within an hour, and a very few taking more than 10 hours (this excludes the time needed for encoding the problem to COP, as we did not optimise that step).
In contrast, inference with MLN takes substantially longer time and was the most time-consuming part of the experiments.
Moreover, the best result on each dataset (Figure \ref{fig:resultsAUCPR}) is rarely achieved with the latent representation with the most expressive \alp{}, which are the runs that take the longest. %\tias{This hyperparam combo is not in the hyperparams described above?}
% (with decoder clause length of 3, and any length of encoder clauses)
%This suggest that the length of clauses serves as a form of regularisation, and one can safely stick to the shorter clauses (and stacking if the complexity is needed) which are fast to learn.

\section{Related Work}
\label{sec:relatedwork}

The most prominent paradigm in merging SRL and DL are (knowledge) graph embeddings~\cite{NickelReview,DBLP:journals/debu/HamiltonYL17}.
In contrast to \alp{s}, these methods do not retain full relational data representation but approximate it by vectorisation.
Several works \cite{DBLP:conf/uai/MinerviniDRR17,demeester2016lifted} impose logical constraints on embeddings but do not retain the relational representation.

\citeauthor{Kazemi2017RelNNAD} [\citeyear{Kazemi2017RelNNAD}] and \citeauthor{LiftedRNN} [\citeyear{LiftedRNN}] introduce symbolic variants of neural networks for relational data.
\citeauthor{Evans2018LearningER} [\citeyear{Evans2018LearningER}] introduce a differentiable way to learn predictive logic programs.
These are likewise capable of discovering latent concepts (predicates), but focus on predictive learning, often with a pre-specified architecture.

Several works integrate neural and symbolic components but do not explore learning new symbolic representation.
\citeauthor{DTP2017} [\citeyear{DTP2017}] introduce a differentiable version of Prolog's theorem proving procedure, which \citeauthor{Campero18} [\citeyear{Campero18}] leverage to acquire logical theories from data.
\citeauthor{NIPS2018_7632} [\citeyear{NIPS2018_7632}] combine symbolic and neural reasoning into a joint framework, but only consider the problem of parameter learning not the (generative) structure learning.
%Hence, they cannot learn the structure of a generative model.
%\tias{And we do because... e.g. 'hence can not learn generative models'?}

Inventing a new relational vocabulary defined in terms of the provided one is known as \textit{predicate invention} in SRL~\cite{Kramer1995,Cropper2018}.
In contrast to \alp{s}, these methods create latent concepts in a weakly supervised manner -- there is no direct supervision for the latent predicate, but there is indirect supervision provided by the accuracy of the predictions.
An exception to this is the work by \citeauthor{Kok:2007:SPI:1273496.1273551} [\citeyear{Kok:2007:SPI:1273496.1273551}]; however, it does not provide novel language constructs to an SRL model, but only compresses the existing data by identifying entities that are \textit{identical}.

We draw inspiration from program induction and synthesis~\cite{Gulwani2017}, in particular, unsupervised methods for program induction~\cite{Ellis:2015:ULP:2969239.2969348,Lake2015HumanlevelCL}.
These methods encode program induction as a constraint satisfaction problem similar to \alp{s}, however, they do not create new latent concepts.

\section{Conclusion}

This work introduce \textit{Auto-encoding Logic Programs} (\alp{s}) -- a novel logic-based representation learning framework for relational data.
The novelty of the proposed framework is that it learns a latent representation in a symbolic, instead of a gradient-based way.
It achieves that by relying on first-order logic as a data representation language, which has a benefit of exactly representing the rich relational data without the need to approximate it in the embeddings spaces like many of the related works.
We further show that learning \alp{s} can be cast as a constraint optimisation problem, which can be solved efficiently in many cases.
We experimentally evaluate our approach and show that learning generative models from the relational latent representations created by \alp{s} results in substantially improved AUC-PR scores compared to learning from the original data representation.

This work shows the potential of latent representations for the SRL community and opens challenges for bringing these ideas to their maturity; in particular, the understanding of the desirable properties of relational representations and the development of scalable methods to create them.

\section*{Acknowledgments}
 \noindent The authors are grateful to Oliver Schulte for his comments on the early version of this work. This work was partially funded by the VLAIO-SBO project HYMOP (150033).

%% The file named.bst is a bibliography style file for BibTeX 0.99c
\bibliographystyle{named}
\bibliography{biblio}

\appendix

\section{Formal Definitions}

Here we provide the formal definitions concerning the introduced \alp{} framework, covering the main components (Section \ref{app:Alp}), incorporation of background knowledge (Section \ref{app:back}) and pruning criteria (Section \ref{app:prune})

\subsection{Auto-encoding Logic Programs}
\label{app:Alp}

\begin{definition}
\textbf{Relational encoder.}
A \textit{relational encoder} is a logic program $\mathcal{E}$ %$: \mathcal{HB(P,C)} \rightarrow \mathcal{HB(L,C)}$
that maps a set of facts $\mathcal{KB} \subset \mathcal{HB(P,C)}$ to a set of latent facts $\mathcal{KB}_{\mathcal{L}} \subset \mathcal{HB(L,C)}$.
The clauses of the encoder are termed \textit{encoder clauses} and their bodies consist of predicates in $\mathcal{P}$, while the heads are composed of predicates in $\mathcal{L}$.
\end{definition}

\begin{definition}
\textbf{Relational decoder.}
A \textit{relational decoder}  is a logic program $\mathcal{D}$ %$: \mathcal{HB(L,C)} \rightarrow \mathcal{HB(P,C)}$
that maps a set of latent facts $\mathcal{KB}_{\mathcal{L}} \subset \mathcal{HB(L,C)}$  to a new set of facts $\mathcal{KB}' \subset \mathcal{HB(P,C)}$.
The clauses are termed \textit{decoder clauses} and their bodies consist of predicates in $\mathcal{L}$, while the heads are predicates in $\mathcal{P}$.
\end{definition}

\begin{definition}
\textbf{Auto-encoding logic program (\alp{}).}
An \textit{auto-encoding logic program} is a logic program that, given a knowledge base $\mathcal{KB}$, constructs encoder $\mathcal{E}$ and decoder $\mathcal{D}$ programs, together with the latent predicate vocabulary $\mathcal{L}$.
\end{definition}

\subsection{Background knowledge}
\label{app:back}

Many SRL systems allow users to provide \textit{background knowledge}: additional knowledge, separate from data, that a learner can leverage to express more complex rules. 
This knowledge typically consists of clauses.
Incorporating such knowledge in \alp{s} is straightforward: predicates and facts provided as background knowledge can be used to construct encoder clauses, but are ignored for reconstruction.

\subsection{Recursion in \alp{s}}

The formalisation of \alp{s} presented in this work defines both encoder and decoder as non-recursive logic programs. 
Incorporating recursion in both encoder and decoder programs is theoretically straightforward:

\begin{definition}
\textbf{Relational encoder.}
A \textit{relational encoder} is a logic program $\mathcal{E}$ %$: \mathcal{HB(P,C)} \rightarrow \mathcal{HB(L,C)}$
that maps a set of facts $\mathcal{KB} \subset \mathcal{HB(P,C)}$ to a set of latent facts $\mathcal{KB}_{\mathcal{L}} \subset \mathcal{HB(L,C)}$.
The clauses of the encoder are termed \textit{encoder clauses} and their bodies consist of predicates in $\mathcal{P} \cup \mathcal{L}$, while the heads are composed of predicates in $\mathcal{L}$.
\end{definition}

\begin{definition}
\textbf{Relational decoder.}
A \textit{relational decoder}  is a logic program $\mathcal{D}$ %$: \mathcal{HB(L,C)} \rightarrow \mathcal{HB(P,C)}$
that maps a set of latent facts $\mathcal{KB}_{\mathcal{L}} \subset \mathcal{HB(L,C)}$  to a new set of facts $\mathcal{KB}' \subset \mathcal{HB(P,C)}$.
The clauses are termed \textit{decoder clauses} and their bodies consist of predicates in $\mathcal{L} \cup \mathcal{P}$, while the heads are predicates in $\mathcal{P}$.
\end{definition}

However, learning recursive \alp{s} through constraint satisfaction is more challenging: one would have to make sure that for every recursive clause, a base (non-recursive) case is also a part of the solution.

\subsection{Pruning the candidates}
\label{app:prune}

\begin{definition}
\textbf{Naming variants.}
Given two encoder clauses \textit{ec$_1$(X):\!-...}
%p1(X,Y),p2(Y)}
and \textit{ec$_2$(X):\!-...}
% p3(X,Y),p2(Y).}
with \textit{L1=\{ec$_1$(a),ec$_1$(b),...\}} and \textit{L2=\{ec$_2$(a),ec$_2$(b), ...\}} the respective sets of true instantiations.
The two clauses are \textit{naming variants} if renaming the predicate names in \textit{L1} and \textit{L2} to a common name yields identical sets \textit{L1} and \textit{L2}.
\end{definition}

\begin{definition}
\textbf{Signature variants.}
Assume two decoder clauses, \texttt{dc}$_m$ and \texttt{dc}$_n$, with the same head predicate.
Let \texttt{C}$_m$ (\texttt{C}$_n$) be the maximal set of facts decoder clauses \texttt{dc}$_m$ (\texttt{dc}$_n$) entail, and \texttt{B}$_m$ (\texttt{B}$_n$) be the maximal set of predicates used in the body of the two clauses.
The two clauses are \textit{signature variants} iff \texttt{C}$_m$ $=$ \texttt{C}$_n$ and \texttt{B$_m$} $=$ \texttt{B$_n$}.
\end{definition}

\begin{definition}
\textbf{Corruption level.}
Given a decoder clause \texttt{dc}$_i$, its true instantiations \texttt{DC} and the original interpretation $\mathcal{KB}$, the \textit{corruption level} is defined as $c(\text{\texttt{dc}}_i) = \frac{|e \in \texttt{DC}\ \wedge\ e \not \in \mathcal{KB}|}{|\text{\texttt{DC}}|}$.
\end{definition}

\section{Constraints}

\paragraph{Connecting encoder and decoder.}
We impose the following constraints connecting the encoder and decoder clauses:
\begin{center}

    \texttt{l}$_i$ $\Leftrightarrow$ \texttt{dc$_m$ $\vee$ dc$_n$}
\end{center}
The constraint states that an encoder clause \texttt{l}$_i$ defining the latent predicate $l$ must be selected if at least one of the decoder clauses using $l$ in the body, \texttt{dc$_m$} and \texttt{dc$_n$}, is selected.

\paragraph{Generality of clauses.}
Assume that the clause $c_1$ is more general than the clause $c_2$.
As $c_2$ cannot bring new information if $c_1$ is also a part of the solution, we impose the constraint stating the not both of the clauses can be selected at the same time:

\begin{center}
   !(\texttt{c}$_1$ $\wedge$ \texttt{c}$_2$) == \texttt{true}.
\end{center}

% \textbf{Syntactic variants} Assuming that \texttt{dc}$_1$,\texttt{dc}$_2$, \texttt{dc}$_3$ are syntactic variants, we impose the constraint of the following form:

% \begin{center}
% 	\texttt{dc}$_1$ + \texttt{dc}$_2$ + \texttt{dc}$_3$ $\leq$ \texttt{1}.
% \end{center}

\paragraph{Reconstruct all predicates.}
Assume that \texttt{dc}$_k$,\texttt{dc}$_l$ and \texttt{dc}$_m$ are decoder clauses all having predicate $p$ as the head predicate.
we introduce the following constraint to state that at least one of them has to be selected: %\tias{the ==1 is highly unconventional, just leave it out? or at least == true like you did earlier}
$$ \text{\texttt{dc}}_1 \vee \text{\texttt{dc}}_2 \vee \text{\texttt{dc}}_3 == \text{\texttt{true}}.$$

\section{Candidate clause generation}

SRL systems rely on \textit{language bias} to construct the space of candidate clauses.
Language bias contains syntactic instructions how to compose predicates to form clauses.
These instructions can often be quite extensive and require an extensive amount of time from the user to specify them correctly.

We employ a two-step approach for enumerating candidate clauses.
The first step \textit{constructs all possible bodies that can be used to formulate a clause.}
To enumerate the bodies, we employ a simple version of the bias based on the argument binding: each argument of the predicates needs to be annotated as either \textit{bounded} ('+') or \textit{unbounded} ('-').
These annotations are used when extending a set of atoms in the body with a new atom: a \textit{bounded} argument of the predicate of the new atom needs to be replaced with an existing variable, whereas \textit{unbounded} argument introduces a new variable.
Consider predicates \texttt{p/2} and \texttt{q/1} with the annotations \texttt{p(}\textit{bound,unbound}\texttt{)} and \texttt{q(}\textit{unbound}\texttt{)}.
The initial set of bodies consists only of the predicates itself
\begin{center}
	\texttt{p(X,Y)} \quad \quad \quad \texttt{q(X)}
\end{center}
Extending \texttt{p(X,Y)} would result in
\begin{center}
	\texttt{p(X,Y),p(Y,Z)} \quad \quad \texttt{p(X,Y),p(X,Z)} \\
	\texttt{P(X,Y),q(X)} \quad \quad \texttt{p(X,Y),q(Y)}.
\end{center}

The second step turns the bodies into the candidate clauses by determining which variables should go to the head of the clauses.
We do this by limiting the number of variables in the head to 2 and creating clauses with all possible combinations of variables if there are more than 2 of them.
For instance, turning \texttt{p(X,Y),p(Y,Z)} into a clause would give three clauses
\begin{center}
	\texttt{h$_1$(X,Y) :- p(X,Y),p(Y,Z)}
	
	\texttt{h$_2$(X,Z) :- p(X,Y),p(Y,Z)}
	
	\texttt{h$_3$(Y,Z) :- p(X,Y),p(Y,Z)}
\end{center}

\section{Comment on alternatives for learning encoder/decoder logic programs}

Instead of learning \alp{s} as a COP problem, one could try to learn encoder and decoder with the existing Inductive logic programming \cite{Muggleton94inductivelogic} learners (such as Aleph).
However, ground truth for the latent predicates is not available -- the goal is to obtain them. 
Moreover, these methods do not allow for constraints on latent representations, something that COP allows us to do naturally.

\section{Experimental details}

\subsection{Adding the decoder to MLN}

In order to allows an MLN model to learn in the latent space but \textit{reason} about data in the original observed space, we append the decoder to the latent MLN model.
An issue with doing that is that the Alps as specified under the \textit{clausal logic} semantics  while MLNs operate under the \textit{first-order logic} semantics.
To convert between the two semantics, we rely on the clausal normal form~\cite{VanEmden:1976:SPL:321978.321991,Russell:2009:AIM:1671238,Jackson:2004:CFC:2103144.2103160} in which the clause
\begin{center}
	\texttt{h($\cdot$) :- b$_1$($\cdot$),b$_2$($\cdot$).}
\end{center}
can be written as the following first-order logic formula
\begin{center}
	\texttt{h($\cdot$) $\vee$ $\neg$b$_1$($\cdot$) $\vee$ $\neg$b$_2$($\cdot$).}
\end{center}

Therefore, we convert each decoder clause in the same way and add it to the MLN as a deterministic formula.

\subsection{On the need for open-world interpretation during evaluation}

In Section \ref{sec:expr} (Practical considerations), we have emphasised the need for declaring some latent predicates as \textit{open-world}. This might seems in conflict with the \textit{closed-world} assumption  \alp{s} inherit from the logic programming framework.
However, the need for open-world assumption is the relict of the evaluation procedure, not a part of the framework.

During evaluation, we select one predicate as a test predicate and use the associated ground atoms (and their truth evaluations) as queries conditioned on the rest of the data.
Consider \textit{p/2} to be the test predicates and the following encoder clause

\begin{center}
	\texttt{latent(X,Y) :- p(X,Y)}.
\end{center} 

As the test predicate is removed from the data, there would be no true instantiations of \texttt{latent/2} if declared closed-world.
Consequently, we would not be able to infer anything using any rule containing \texttt{latent/2}.
To prevent this, we declare \texttt{latent/2} to be an open-world predicate so that the reasoning engine has to deduce the truth value of its instantiations.

\subsection{Performances of the validation test}

\begin{figure*}
	\centering
	\begin{subfigure}{.4\linewidth}
		\centering
		\includegraphics[width=.7\linewidth]{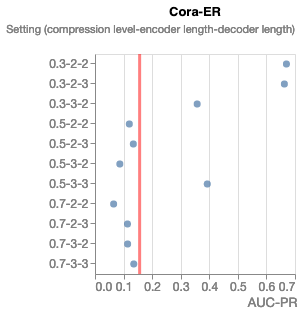}
		\caption{Cora-ER}
	\end{subfigure}
	~
	\begin{subfigure}{.4\linewidth}
		\centering
		\includegraphics[width=.7\linewidth]{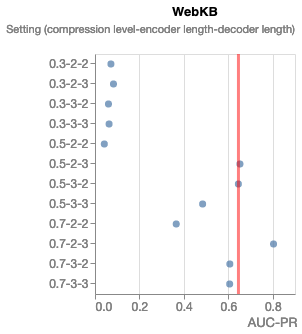}
		\caption{WebKB}
	\end{subfigure}
	~
	\begin{subfigure}{.4\linewidth}
		\centering
		\includegraphics[width=.7\linewidth]{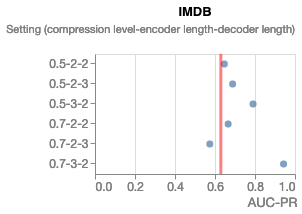}
		\caption{IMDB}
	\end{subfigure}
	~
	\begin{subfigure}{.4\linewidth}
		\centering
		\includegraphics[width=.7\linewidth]{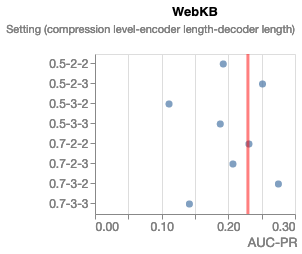}
		\caption{UWCSE}
	\end{subfigure}
	\caption{Performance of the baseline and the latent MLN on the validation set. The baseline MLN is indicated with a red line, while the performances of the latent MLNs (different versions correspond to the parameters of the latent representation) are indicated with dots. Unreported combination of Alp hyper-parameters indicate that it was not possible to learn the latent representation. \label{fig:validation}}
\end{figure*}

The performance of the MLN models on the validation set are reported in Figure \ref{fig:validation}. The models are learned on the original data representation (red line) and latent representations created by imposing different values on the hyper--parameters of Alps.

\subsection{Reconstruction vs AUC-PR}

\begin{figure}
	\centering
	\includegraphics[width=.7\linewidth]{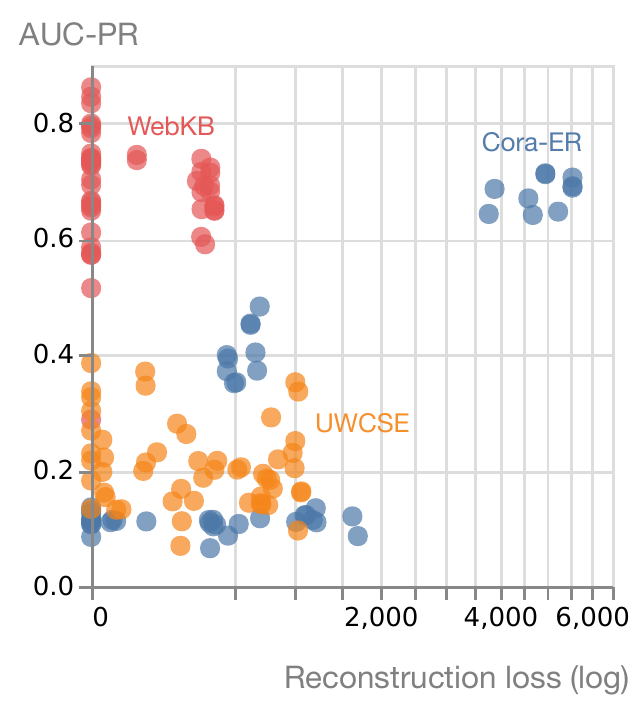}
	\caption{The relationship between reconstruction loss and AUC-PR.} \label{fig:rlvsaucpr}
\end{figure}

Figure \ref{fig:rlvsaucpr} illustrates the relationship between AUC-PR and reconstruction loss. 
The results illustrate  that better reconstruction loss does not necessarily lead to better performance.
That is particularly evident in the case of Cora-ER where the best performing models correspond to the worst reconstruction losses.
This indicates that the non-reconstructed facts are either noisy or irrelevant and consequently make the subsequent learning easier when removed from the data.

\end{document}